\documentclass[journal]{IEEEtran}

\ifCLASSINFOpdf
\else
   \usepackage[dvips]{graphicx}
\fi
\usepackage{url}

\usepackage{amsmath}
\usepackage{indentfirst}
\usepackage{amssymb}
\usepackage{subfigure}
\usepackage{algorithm}
\usepackage{algorithmic}
\usepackage{graphicx}
\usepackage{multirow}
\usepackage{hyperref}
\usepackage{times}
\usepackage{epsfig}
\usepackage{subfigure}
\usepackage{graphicx}
\usepackage{amsmath}
\usepackage{amssymb}
\usepackage{multirow}
\usepackage{booktabs}       
\usepackage{epstopdf}
\usepackage{verbatim}
\begin{document}

\title{Learning Diverse Features with Part-Level Resolution for Person Re-Identification}
\author{Ben Xie, Xiaofu Wu$^\dag$, Suofei Zhang, Shiliang Zhao and Ming Li
\thanks{$^\dag$Corresponding author. This work was supported in part by the National Natural Science Foundation of China under Grants 61372123, 61671253 and by the Scientific Research Foundation of Nanjing University of Posts and Telecommunications under Grant NY213002.}
\thanks{Ben~Xie, Xiaofu~Wu and Shiliang~Zhao are with the National Engineering Research Center of Communications and Networking, Nanjing University of Posts and Telecommunications, Nanjing 210003, China (E-mails: 1018010631@njupt.edu.cn; xfuwu@ieee.org; 1018010632@njupt.edu.cn).}
\thanks{Suofei Zhang is with the School of Internet of Things, Nanjing University of Posts and Telecommunications, Nanjing 210003, China (E-mail: zhangsuofei@njupt.edu.cn).}
\thanks{Ming~Li is with the Supply Chain Platform Division, Alibaba Group, Hangzhou 311121, China
(Email: sebastian.lm@alibaba-inc.com).}}

\maketitle

\begin{abstract}
   Learning diverse features is key to the success of person re-identification. Various part-based methods have been extensively proposed for learning local representations, which, however, are still inferior to the best-performing methods for person re-identification. This paper proposes to construct a strong lightweight network architecture, termed PLR-OSNet, based on the idea of Part-Level feature Resolution over the Omni-Scale Network (OSNet) for achieving feature diversity. The proposed PLR-OSNet has two branches, one branch for global feature representation and the other branch for local feature representation. The local branch employs a uniform partition strategy for part-level feature resolution but produces only a single identity-prediction loss, which is in sharp contrast to the existing part-based methods.  Empirical evidence demonstrates that the proposed PLR-OSNet achieves state-of-the-art performance on popular person Re-ID datasets, including Market1501, DukeMTMC-reID and CUHK03, despite its small model size.
\end{abstract}


\IEEEpeerreviewmaketitle

\newtheorem{plm}{Problem}
\newtheorem{thm}{Theorem}
\section{Introduction}
In recent years, person re-identification (Re-ID) has attracted increasing interest due to its fundamental role in emerging computer vision applications such as video surveillance, human identity validation, and authentication, and human-robot interaction \cite{zheng2016person,dai2019BDB,chen2019MHN,yang2019CAMA,hou2019IAN,chen2019ABD}. The objective of person Re-ID is to match any query image with the images of the same person taken by the same or different cameras at different angles, time or location. Despite its recent progress, identifying the person of interest accurately and reliably is still very challenging due to  huge variations in lighting, human pose, background, camera viewpoint, etc. With the ID-labeled training set, one of the main goals in the field of person Re-ID is to discover a low-dimensional but rich representation of any input image for person matching.
\begin{figure}[t]
\centering
\subfigure{
\includegraphics[width=8.0cm]{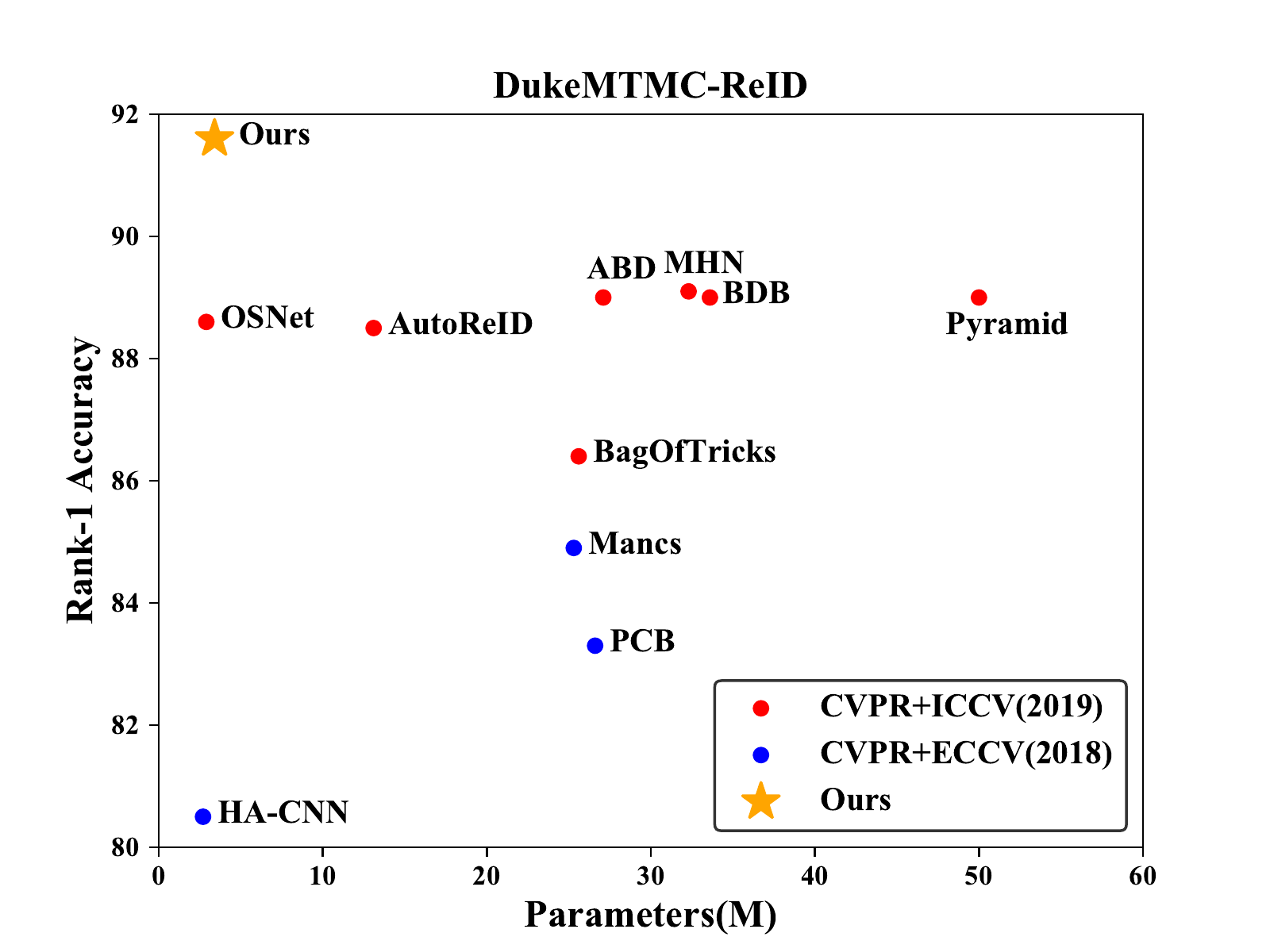}}
\subfigure{
\includegraphics[width=8.0cm]{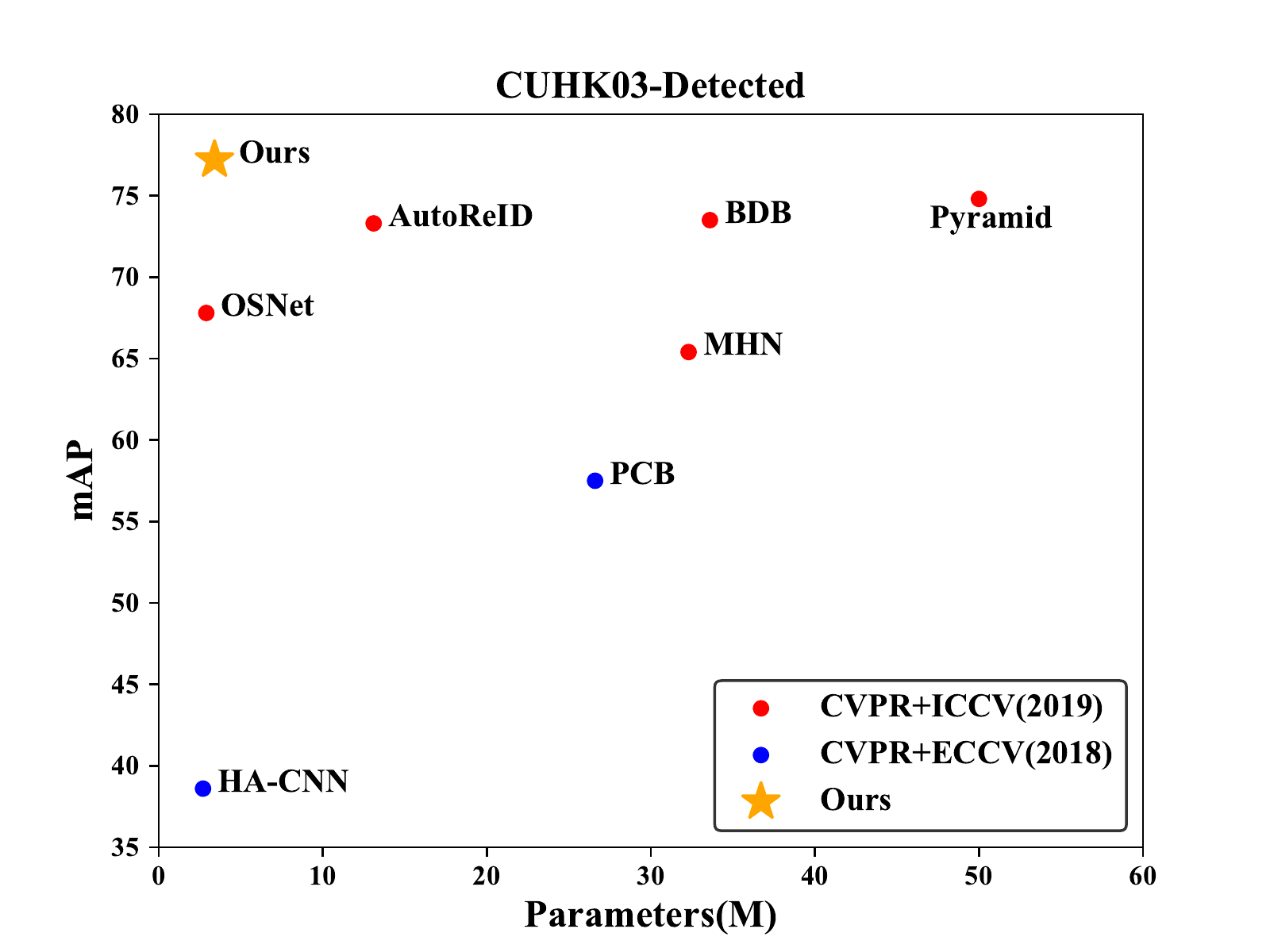}}
\caption{The performance of different baselines on the DukeMTMC-reID and CUHK03-Detected datasets. We compare the proposed method with other baselines published in CVPR, ECCV and ICCV (2018/2019).}
\label{fig:dem}
\end{figure}


Person Re-ID was often formulated as a metric-learning problem (or a feature-embedding problem) ~\cite{su2017pose,chen2017beyond,bai2017reid}, where the distance between intra-class samples is required to be less than the distance between inter-class ones by at least a margin. Unfortunately, a direct implementation of this idea requires to group samples in a pairwise manner, which is known to be computationally intensive. Alternatively, a classification task is employed to find the feature-embedding solution due to its advantage on the implementation complexity. Currently, various state-of-the-art methods \cite{he2019Bag,dai2019BDB,chen2019ABD,zheng2017discriminatively,hermans2017defense} for person Re-ID have evolved from a single metric-learning problem or a single discriminative classification problem to a multi-task problem, where both the discriminative loss and the triplet loss are employed~\cite{su2016deep}. As each sample image is only labeled with the person ID, an end-to-end training approach usually has difficulty to learn diverse and rich features without elaborate design of the underlying neural network and further use of some regularization techniques.

In the past years, various part-based approaches~\cite{yao2017deep,sun2018beyond,zhao2017deeply} and dropout-based approaches~\cite{dai2018batch} have been proposed in order to learn rich features from the ID-labeled dataset. Differing from conventional pose-based Re-ID approaches~\cite{su2017pose,kumar2017pose,zheng2017pose,qian2018pose}, part-based approaches usually locate a number of body parts firstly, and force each part meeting an individual ID-prediction loss in getting discriminative part-level feature representations \cite{wang2018MGN,suh2018part,cheng2016person,fan2018scpnet}.
Dropout-based approaches, however, intend to discover rich features from enlarging the dataset with various dropout-based data-augmentation methods, such as cutout \cite{DeVries2017Cut} and random erasing \cite{zhong2017Erasing}, or from dropping the intermediate features from feature-extracting networks, such as Batch BropBlock \cite{dai2019BDB}.

The performance of part-based methods relies heavily on the employed partition mechanism. Semantic partitions may offer stable cues to good alignment but are prone to noisy pose detections, as it requires that human body parts should be accurately identified and located.  The uniform horizontal partition was widely employed in \cite{sun2018beyond,suh2018part}, which, however, provides limited performance improvement.

This motivates the work in this paper, where we propose a novel two-branch lightweight architecture for discovering rich features in person Re-ID. In particular, we employ the idea of part-level feature resolution in developing a strong two-branch baseline for person Re-ID. Compared to the popular part-based method of PCB \cite{sun2018beyond}, our method differs mainly in two aspects. One is the use of global branch for facilitating the extraction of a global feature, and the other is the use of a single ID-prediction loss for part-level feature resolution. We briefly summarize the main contribution of this paper as follows:
\begin{enumerate}

\item
Based on the omni-scale network (OSNet) baseline \cite{zhou2019OSNet}, we propose a lightweight two-branch network architecture (PLR-OSNet) for person Re-ID. Its global branch adopts a global-max-pooling layer while its local branch employs a part-level feature resolution scheme for producing only a single ID-prediction loss, which is in sharp contrast to existing part-based methods. The proposed architecture is shown to be effective for achieving feature diversity.

\item
  Despite its small model size, the proposed PLR-OSNet is very efficient as depicted in Figure \ref{fig:dem} for achieving the state-of-the-art results \cite{li2018harmonious,wang2018mancs,chen2019ABD,chen2019MHN,dai2019BDB,he2019Bag,quan2019Auto,sun2018beyond,zheng2019pyramid} on the three popular person Re-ID datasets, Marktet1501, DukeMTMC-reID and CUHK03. It achieves the rank-1 accuracy of 91.6\% for DukeMTMC-reID and 83.5\% for CUHK03-Labeled without using re-ranking. \footnote{Source codes are available at \href{url}{https://github.com/AI-NERC-NUPT/PLR-OSNet}}

\end{enumerate}

\section{Related Work}
\label{gen_inst}
We review the relevant work about embedded feature learning, discriminative feature learning, part-based
feature learning, and multi-scale feature learning. Besides of various problems for person Re-ID, we are particularly interested in achieving feature diversity throughout this section.

\subsection{Embedded Feature Learning}
Person Re-ID can be formulated as a feature-embedding problem, which looks for a function mapping the high-dimensional pedestrian images into a low-dimensional feature space. This feature-embedding formulation requires the mapping function to ensure that given any anchor image, any positive image from the same person should has a lower distance to the anchor in the feature space compared to any negative image from a different person. This is known to be the objective of the triplet loss in training.

For an efficient feature-embedding learning, the batch hard triplet loss \cite{Ristani2018Soft} was proposed to mine the hardest positive and the hardest negative samples for each pedestrian image in a batch. However, it is sensitive to outlier samples and may discard useful information due to its hard selective approach. To deal with these problems, Ristani et al. proposed the batch-soft triplet loss \cite{Ristani2018Soft}, which introduce a weighting factor for each pair distance.  One hyper-parameter that exists in all of the triplet loss variations is the margin. To eliminate the manual parameter of the margin,  the softplus function $\ln(1 +\exp(\cdot))$ instead of $[\cdot]_+ = \max(0, \cdot)$ in the triplet loss function was introduced in \cite{hermans2017defense}, which is known to be soft-margin triplet loss.

\subsection{Discriminative Feature Learning}
A feature-embedding approach requires to group the pedestrian images into pairs for training, which is not efficient in general. Discriminative feature learning is more efficient by training a classification task, where each person is regarded as a single class. As each pedestrian image in a training dataset is only labeled by a single person ID, a fundamental problem in the field of person Re-ID is how to learn diverse features from the ID-labeled dataset.

To get diverse features from an end-to-end training approach, multi-branch network architectures have been widely employed~\cite{sun2018beyond,dai2019BDB,chen2017multi}, where a shared-net is often followed by multiple subnetwork branches. To achieve feature diversity, distinct mechanisms should be imposed among different branches, such as attention \cite{chen2019ABD,chen2019MHN}, feature dropping \cite{dai2019BDB,xia2019SONA}, and overlapped activation penalty \cite{yang2019CAMA} .
\begin{figure*}
\begin{center}
\includegraphics[width=0.95\textwidth]{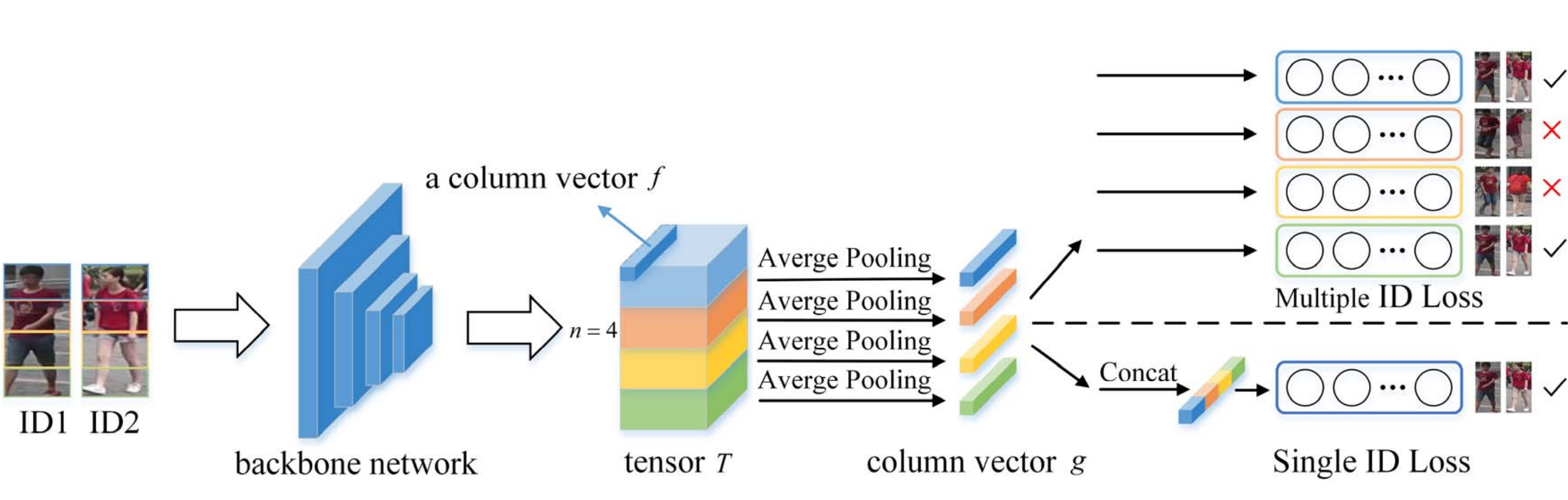} 
\end{center}
   \caption{Single ID loss vs. multiple ID loss with part-level feature resolution. The input image goes through backbone network to obtain a 3-D tensor, which is vertically split into $n$ = 4 part-level features, and then averages each part-level tensor into a vector. These $n$ part-level vectors are used to drive $n$ independent ID losses in PCB, or simply concatenated for driving a single ID loss in training. The use of multiple ID loss in training may lead to false prediction of the person ID with some part-level features.}
\label{fig:PLR}
\end{figure*}

\subsection{Part-Based Feature Learning}
Part-based feature learning with hand-crafted algorithms had been pursued for a long time for the purpose of person retrieval before the era of deep-learning. In \cite{sun2018beyond}, the Part-based Convolutional Baseline (PCB) network was proposed, which employs uniform partition on the conv-layer for learning part-level features. Essentially, it employed a 6-branch network by dividing the whole body into 6 horizontal stripes in the feature space and each part feature vector was used to produce an independent ID-prediction loss. The idea of PCB was very welcome and widely adopted for developing stronger methods in the recent years for person Re-ID \cite{zheng2019pyramid}\cite{quan2019Auto}\cite{wang2018MGN}.

The part-level feature learning has an intuitive advantage for extracting diverse features from the ID-labeled pedestrian images. However, the pristine division strategy usually suffers from misalignment \cite{zhang2020Part} between corresponding parts due to large variations in poses, viewpoints and scales. In particular, the use of multiple ID-prediction loss (an independent ID loss for each part) may fail to capture the semantic part-level features since a pedestrian image may simply contain the semantically different parts at a uniformly-divided pedestrian part. This may partially explain the limited performance advantage of various PCB-based algorithms, compared to the state-of-the-art methods \cite{xia2019SONA,chen2019MHN}.

\subsection{Multi-Scale Feature Learning}
Recently, multi-scale feature learning, together with the multi-stream building block design \cite{chang2018MFN,qian2017mulscale,chen2018person}, has been proved to be efficient for improving the performance of person Re-ID. By designing a residual block composed of multiple convolutional feature streams, each detecting features at a certain scale, the concept of omni-scale deep feature learning was further introduced in \cite{zhou2019OSNet} and a lightweight CNN architecture, termed OSNet, was cleverly constructed for learning omni-scale feature representations. Experiments demonstrated that OSNet performs very well for both tasks of classification and person Re-ID, despite its lightweight design.

\section{PLR-OSNet}
\label{headings}
\subsection{Part-Level Feature Resolution}
\begin{figure*}
\begin{center}
\includegraphics[width=0.95\textwidth]{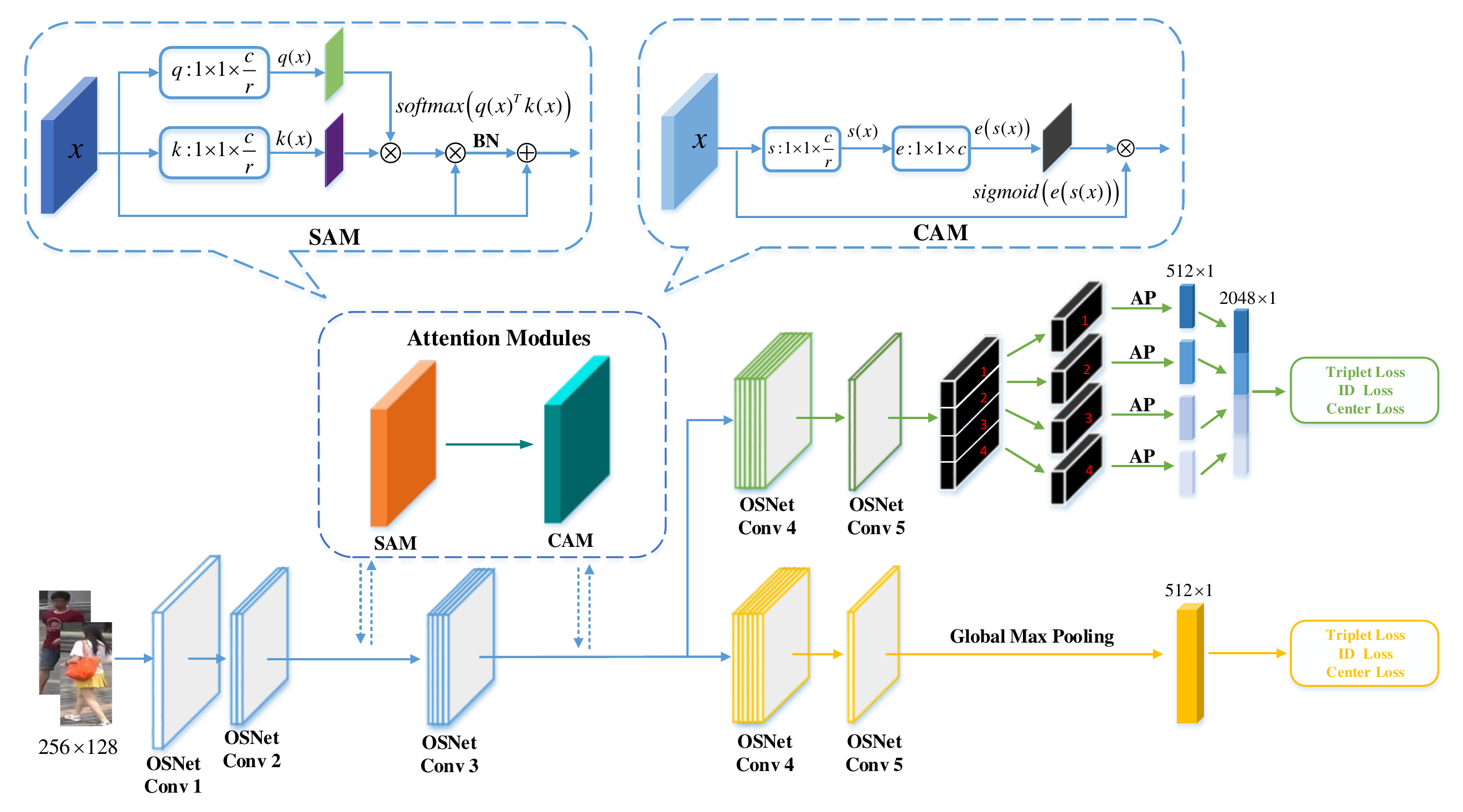} 
\end{center}
   \caption{The overall network architecture of PLR-OSNet. During testing, the feature embedding concatenated from both global branch and local branch is used for the final matching distance computation.}
\label{fig:net}
\end{figure*}
For PCB with $n$-part resolution, it produces $n$ part-level feature vectors by dividing the whole body into $n$ horizontal stripes in the feature space. As shown in Figure \ref{fig:PLR}, the input image goes forward through the stacked convolutional layers from the backbone network to form a 3-D tensor $T$. PCB employs a conventional average pooling layer to spatially down-sample $T$ into $n$ pieces of column vectors $\mathbf{g}_1,\mathbf{g}_2, \cdots,\mathbf{g}_n$, followed by $n$ classifiers in order to produce $n$ ID-predication loss. Note that the classifier is implemented by a fully-connected (FC) layer and a softmax function. Hence, when the batch of input labeled samples are $ \{(x_i,y_i), i=1,\cdots, N_s\}$, PCB employs the multiple ID-prediction loss as
\begin{eqnarray}
L_{m-id}  &=& \sum_{p=1}^n L_{id}^p, \\
L_{id}^p &=& - \frac{1}{N_s}\sum_{i=1}^{N_s} \log \left(\frac{\exp((\mathbf{W}_p^{y_i})^T \mathbf{g}_p^i + b_{y_i})}{\sum_j\exp((\mathbf{W}_p^{j})^T \mathbf{g}_p^j + b_j)}\right).
\end{eqnarray}
where $\mathbf{W}_p^j$, $\mathbf{W}_p^{y_i}$ are the $j$-th and $y_i$-th column of the weight matrix $\mathbf{W}_p$ (the $p$-th classifier designated for $\mathbf{g}_p$), respectively.

By forcing each part-level feature vector to meet an independent ID-prediction loss, one may obtain useful part-level features for discriminating different persons. However, many part-level feature vectors may simply fail to catch any discriminative information for different persons, as shown in Figure \ref{fig:PLR}. Therefore, the use of PCB is practically limited for getting discriminative part-level information.

In order to learn discriminative features with part-level resolution, we propose to concatenate $n$ part-level feature vectors into a single column vector
\begin{equation}
\label{eq:AP}
\mathbf{g}=[\mathbf{g}_1^T,\mathbf{g}_2^T, \cdots,\mathbf{g}_n^T]^T,
\end{equation}
which is further used to produce the ID-prediction loss
\begin{eqnarray}
L_{s-id}  = - \frac{1}{N_s}\sum_{i=1}^{N_s} \log \left(\frac{\exp((\mathbf{W}^{y_i})^T \mathbf{g}^i + b_{y_i})}{\sum_j\exp((\mathbf{W}^{j})^T \mathbf{g}^j + b_j)}\right).
\end{eqnarray}
Here, $\mathbf{W}^j$, $\mathbf{W}^{y_i}$ are the $j$-th and $y_i$-th column of the weight matrix $\mathbf{W}$ (the single  classifier for $\mathbf{g}$), respectively. As the vector $\mathbf{g}$ contains the full information about the input image, the use of a single ID-prediction loss could drive $\mathbf{g}$ to learn sufficient discriminative information.

The proposed approach is somewhat similar to OSNet, where the tensor $T$ is followed by a  global average pooling (GAP) for getting a global descriptor
\begin{equation}
\label{eq:gap}
\mathbf{\bar{g}} = \frac{1}{n}\sum_{p=1}^n \mathbf{g}_p.
\end{equation}
Instead of using the GAP in OSNet, the proposed part-level resolution approach uses average pooling in each part to retrieve part-level feature vectors and the final descriptor $\mathbf{g}$ (\ref{eq:AP}) is of rich local information, which might be simply filtered with the GAP (\ref{eq:gap}) in OSNet.

\subsection{Proposed Network Architecture}
We employ an $2$-branch neural network architecture, by modifying the recently-proposed OSNet baseline. Figure \ref{fig:net} shows the overall network architecture, which includes a backbone network, a global branch (orange colored arrows), and a local branch (blue colored arrows).

\subsubsection{Attention Modules}
Compared to the OSNet, attention modules are explicitly employed in Figure \ref{fig:net}, where both spatial attention module (SAM) and channel attention module (CAM) are used in the shared-net.

For SAM, we employ the version of \cite{chen2019ABD}, which was designed to capture and aggregate those semantically related pixels in the spatial domain. To further reduce the computational complexity, we use a $1\times1$ convolution that forms a functions $q(x)$ (or $k(x)$) to reduce the number of channels $c$ to $c/r$ of the input $x$.

For CAM, the squeeze-and-excitation mechanism \cite{hu2018SE} is employed with slight modifications detailed in Figure \ref{fig:net}. Compared to the channel attention module in \cite{chen2019ABD}, it does not require to compute the channel affinity matrix and therefore can be implemented more efficiently.

\subsubsection{Shared-Net}
The recently-proposed OSNet is employed as the backbone network for feature extraction. OSNet uses a lightweight network architecture for omni-scale feature learning, which is achieved by employing the factorised convolutional layer, the omni-scale residual block and the unified aggregation gate. The shared-net consists of  the first 3 conv layers and 2 transition layers from OSNet. As shown in Figure \ref{fig:net}, we insert SAM + CAM  modules in both conv-2 and conv-3 layers for the shared-net.

\subsubsection{Global Branch with Global-Max-Pooling}
The global branch consists of  the conv4 and conv5 layers, a Global-Max-Pooling (GMP) layer to produce a 512-dimensional vector, providing a compact global feature representation for both the triplet loss and the ID-prediction loss. The use of GMP is mainly for achieving the feature diversity between the global branch and the local branch, where average pooling is known to be popular in PCB \cite{sun2018beyond} and adopted in the local branch.

\subsubsection{Local Branch with Part-Level Feature Resolution}
The local branch has the similar layer structure but with the average pooling (AP) in replace of GMP. To achieve feature diversify, a uniform partition strategy is employed for part-level feature resolution, and four 512-dimensional features are then concatenated for producing just one ID-prediction loss. The use of a single ID-prediction loss is unique in this paper, while PCB and its variations employed a multiple ID-prediction loss with an independent ID-prediction loss for each part.

\subsection{Loss Functions}
The feature vectors from the global and local branches are concatenated as the final descriptor for the person
Re-ID task.  The loss function at either the global branch or the local branch is the sum of a single ID loss (softmax loss), a soft margin  triplet loss~\cite{hermans2017defense} and a center loss~\cite{wen2016discriminative}, namely,
\begin{eqnarray}
 L_{total} = L_{s-id} + \gamma_t L_{triplet} + \gamma_c L_{center},
\end{eqnarray}
where $\gamma_t, \gamma_c$ are weighting factors.

\section{Experiments}
Extensive experiments have been performed for evaluating the effectiveness of the proposed approach over three public person Re-ID datasets: Market1501, DukeMTMC-reID and CUHK03. The results are compared to the state-of-the-art methods.

\subsection{Datasets}

The Market1501 dataset  \cite{zheng2015scalable} has 1,501 identities collected by six cameras and a total of 32,668 pedestrian images. Following \cite{zheng2015scalable}. The dataset is split into a training set with 12,936 images of 751 identities and a testing set of 3,368 query images and 15,913 gallery images of 750 identities.

The DukeMTMC-reID dataset \cite{Ristani2016Performance} contains 1,404 identities captured
by more than 2 cameras and a total of 36,411 images. The training subset contains 702 identities with 16,522 images and the testing subset has other 702 identities.

The CUHK03 dataset \cite{Li2014DeepReID} contains labeled 14,096 images and detected 14,097 images of a total of 1,467
identities captured by two camera views. With splitting just like in \cite{zheng2015scalable}, a non-overlapping 767 identities are for training and 700 identities for testing. The labeled dataset contains 7,368 training images, 5,328 gallery, and 1,400 query images for testing, while the detected dataset contains 7,365 images for training, 5,332 gallery, and 1,400 query images for testing.

\subsection{Implementation Details}
Our network is trained using a single Nvidia Tesla P100 GPU with a batch size of 64. Each identity contains 4 instance images in a batch, so there are 16 identities per batch. The backbone OSNet is initialized from the ImageNet pre-trained model. The total number of epoches is set to 120 [150], namely, 120 for both Market-1501 and DukeMTMC-reID,  and 150 for CUHK03, respectively.  We use the Adam optimizer with the base learning rate initialized to 3.5e-5. With a linear warm-up strategy in first 20 [40] epochs, the learning rate increases to 3.5e-4. Then, the learning rate is decayed to 3.5e-5 after 60 [100] epochs, and further decayed to 3.5e-6 after 90 [130] epochs.

For training, the input images are re-sized to $256\times 128$ and then augmented by random horizontal flip, random erasing, and normalization. The testing images are re-sized to $256 \times 128$ with normalization.

\subsection{Comparison with State-of-the-art Methods}

We compare our work with state-of-the-art methods, in particular emphasis on the recent remarkable works (CVPR'19 and ICCV'19) on person Re-ID, over the popular benchmark datasets Market-1501, DukeMTMC-ReID and CUHK03. All reported results are obtained without any re-ranking \cite{zhong2017re,saquib2018pose} or multi-query fusion \cite{zheng2015scalable} techniques. The comparison results are listed in Table 1, Table 2 and Table 3. From these tables, one can observe that our proposed method performs competitively among various state-of-the-art methods, including PCB \cite{sun2018beyond}, IAN \cite{hou2019IAN}, CAMA \cite{yang2019CAMA},MHN \cite{chen2019MHN},Pyramid~\cite{zheng2019pyramid}, BagOfTricks \cite{he2019Bag}, ABD-Net \cite{chen2019ABD}, BDB \cite{dai2019BDB}, SONA \cite{xia2019SONA}, Auto-ReID \cite{quan2019Auto}, OSNet \cite{zhou2019OSNet}, et al.

As shown, \textit{our PLR-OSNet has achieved the best mAP performance among various state-of-the-art methods for all the three datasets}. For DukeMTMC-reID, PLR-OSNet obtained 91.6\% Rank-1 accuracy and 81.2\% mAP, which significantly outperforms all existing methods. For CUHK03, PLR-OSNet even outperforms SONA in both mAP and Rank-1 accuracy, which might be the best performing algorithm for CUHK03.

Besides of its strong competition in both Rank-1 and mAP performance, PLR-OSNet has a lightweight network architecture inherited from OSNet. It only has only 3.4M parameters while the recently-available Robust-ReID has 6.4M parameters.

\begin{table}
\begin{center}
\label{my-label}
\begin{tabular}{l||cc}
\toprule[1.5pt]
Method     & mAP & rank-1  \\\hline
KPM \cite{shen2018KPM}(CVPR'18)       & 75.3 & 90.1   \\
MLFN \cite{chang2018MFN}(CVPR'18)     & 74.4 & 90.0   \\
CRF \cite{chen2018CRF}(CVPR'18)       & 81.6 & 93.5   \\
HA-CNN \cite{li2018harmonious}(CVPR'18) & 75.7 & 91.2   \\
PCB \cite{sun2018beyond}(ECCV'18)       & 81.6 & 93.8  \\
Mancs \cite{wang2018mancs} (ECCV'18)                         & 82.3 & 93.1 \\
SNL \cite{li2018SNL}(ACM'18) & 73.43 & 88.27  \\
HDLF\cite{zeng2018HDLF}(ACM MM'18) & 79.10 & 93.30  \\
MGN \cite{wang2018MGN}(ACM MM'18) & 86.9 & 95.7  \\
Local CNN\cite{yang2018LCNN}(ACM MM'18) & 87.4 & \bf 95.9  \\
IAN \cite{hou2019IAN} (CVPR'19)       & 83.1 & 94.4  \\
CAMA \cite{yang2019CAMA}(CVPR'19)       & 84.5 & 94.7 \\
MHN \cite{chen2019MHN}(CVPR'19)       & 85.0 & 95.1   \\
Pyramid~\cite{zheng2019pyramid}(CVPR'19) & 88.2 & 95.7 \\
BagOfTricks \cite{he2019Bag} (CVPRW'19)                   & 85.9 & 94.5 \\
ABD \cite{chen2019ABD} (ICCV'19)       & 88.28 &95.6  \\
BDB \cite{dai2019BDB} (ICCV'19)       & 86.7 & 95.3  \\
SONA \cite{xia2019SONA} (ICCV'19)       & 88.67 &95.68  \\
Auto-ReID \cite{quan2019Auto} (ICCV''19)                    & 85.1 & 94.5 \\
OSNet \cite{zhou2019OSNet} (ICCV'19)                       & 84.9 & 94.8  \\\hline
PLR-OSNet       & \bf88.9 & 95.6 \\
\bottomrule[1.5pt]
\end{tabular}
\end{center}
\caption{Comparison of our proposed method with state-of-the-art methods for the Market-1501 dataset.}
\label{tb:market1501}
\end{table}

\begin{table}
\begin{center}
\label{my-label}
\begin{tabular}{l||ccc}
\toprule[1.5pt]
Method     & mAP & rank-1   \\\hline
MLFN \cite{chang2018MFN}(CVPR'18)     & 62.8 & 81.2   \\
GP-Re-ID \cite{almazan2018re} (CVPR'18)  & 72.8 & 85.2 \\
HA-CNN \cite{li2018harmonious}(CVPR'18)  & 63.8 & 80.5 \\
PCB \cite{sun2018beyond} (ECCV'18)       & 69.2 & 83.3  \\
Mancs (ECCV'18)                           & 71.8 & 84.9  \\
MGN \cite{wang2018MGN}(ACM MM'18) & 78.40 & 88.7 \\
Local CNN\cite{yang2018LCNN}(ACM MM'18) & 66.04 & 82.23 \\
IAN \cite{hou2019IAN} (CVPR'19)       & 73.4 & 87.1  \\
CAMA \cite{yang2019CAMA} (CVPR'19)       & 72.9 & 85.8 \\
MHN \cite{chen2019MHN} (CVPR'19)       & 77.2 & 89.1  \\
Pyramid~\cite{zheng2019pyramid}(CVPR'19) & 79.0 &89.0 \\
BagOfTricks (CVPRW'19)                   & 76.4 & 86.4 \\
ABD \cite{chen2019ABD} (ICCV'19)       & 78.59 &89.0  \\
BDB \cite{dai2019BDB} (ICCV'19)       & 76.0 & 89.0  \\
SONA \cite{xia2019SONA} (ICCV'19)       & 78.05 &89.25 \\
Auto-ReID \cite{quan2019Auto} (ICCV'19)                     & 75.1 &88.5 \\
OSNet\cite{zhou2019OSNet} (ICCV'19)                                & 73.5 & 88.6 \\ \hline
PLR-OSNet       & \bf81.2 & \bf91.6 \\
\bottomrule[1.5pt]
\end{tabular}
\end{center}
\caption{Comparison of our proposed method with state-of-the-art methods for the DukeMTMC-reID dataset.}
\label{tb:Duke}
\end{table}

\begin{table}
\begin{center}
\begin{tabular}{l|c@{\hskip 5pt}c@{\hskip 5pt}|c@{\hskip 5pt}c@{\hskip 5pt}}
\toprule[1.5pt]
\multirow{2}{*}{Method}	&	\multicolumn{2}{c|}{Labeled}	&			\multicolumn{2}{c}{Detected}			\\
\cline{2-5}
	&	mAP 	&	rank-1	&	mAP 	&	rank-1	\\
\hline
DaRe+RE~\cite{wang2018resource}(CVPR'18)	&	61.6 	&	66.1	&	59.0	&	63.3 \\
MLFN~\cite{chang2018MFN}(CVPR'18)	&	49.2 	&	54.7	&	47.8	&	52.8	\\
HA-CNN~\cite{li2018harmonious}(CVPR'18)	&	41.0 	&	44.4	&	38.6	&	41.7	\\
PCB~\cite{sun2018beyond}(ECCV'18) &    -   &   -   &	57.5	&	63.7	\\
Mancs (ECCV'18)                   & 63.9    & 69.0  &   60.5    &   65.5    \\
MGN~\cite{wang2018MGN}(ACM MM'18)	&	67.4 	&	68.0 	&	66.0 	&	68.0 	\\
MHN \cite{chen2019MHN} (CVPR'19)   &72.4    & 77.2 & 65.4 & 71.7 \\
Pyramid~\cite{zheng2019pyramid}(CVPR'19)	&	76.9	&	78.9	&	74.8	&	78.9	\\
BDB~\cite{dai2019BDB} (ICCV'19)	&	76.7	&	79.4	&	73.5	&	76.4	\\
SONA \cite{xia2019SONA} (ICCV'19)       & 79.23 & 81.85 & 76.35 & 79.10   \\
Auto-ReID \cite{quan2019Auto} (ICCV'19)                     & 73.0  & 77.9  & 69.3 & 73.3 \\
OSNet \cite{zhou2019OSNet} (ICCV'19)                 & --        & --        &   67.8    &   72.3 \\ \hline
PLR-OSNet 	& \bf80.5	&	\bf84.6	& \bf77.2	& \bf80.4	\\
\bottomrule[1.5pt]
\end{tabular}
\end{center}
\caption{Comparison of our proposed method with state-of-the-art methods for the CUHK03 dataset.}
\label{tbl:CUHK03}
\end{table}

\subsection{Visualization}
\begin{figure}[t]
\centering
\includegraphics[width=8.0cm]{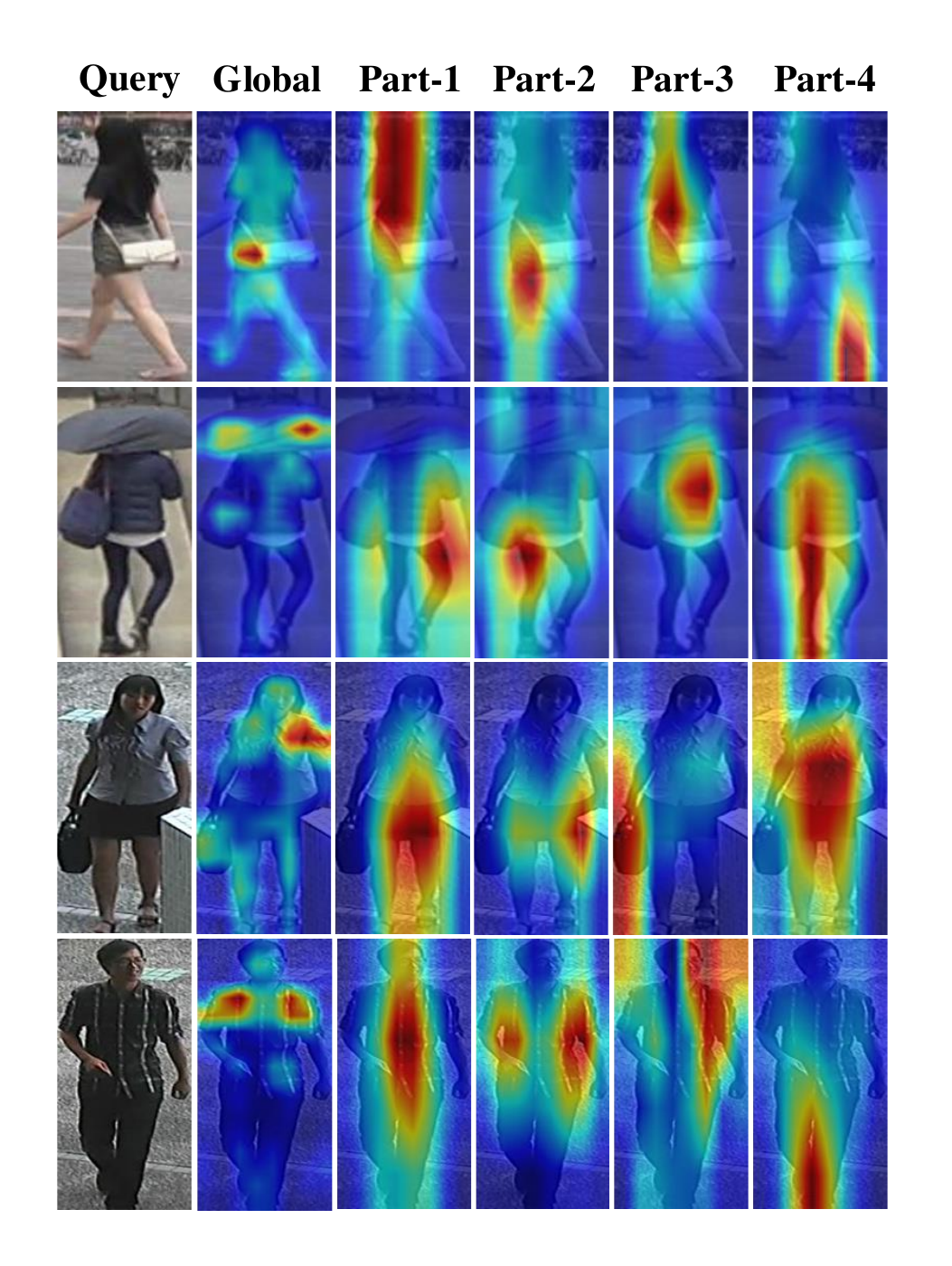}
\caption{Visualization of class activation maps (CAMs) for the global branch and  the local branch (including 4 part-level feature vectors). The proposed architecture allow the model to learn diverse features (marked in orange).}
\label{fig:demDetail}
\end{figure}

\begin{figure}[t]
\centering
\includegraphics[width=8.0cm]{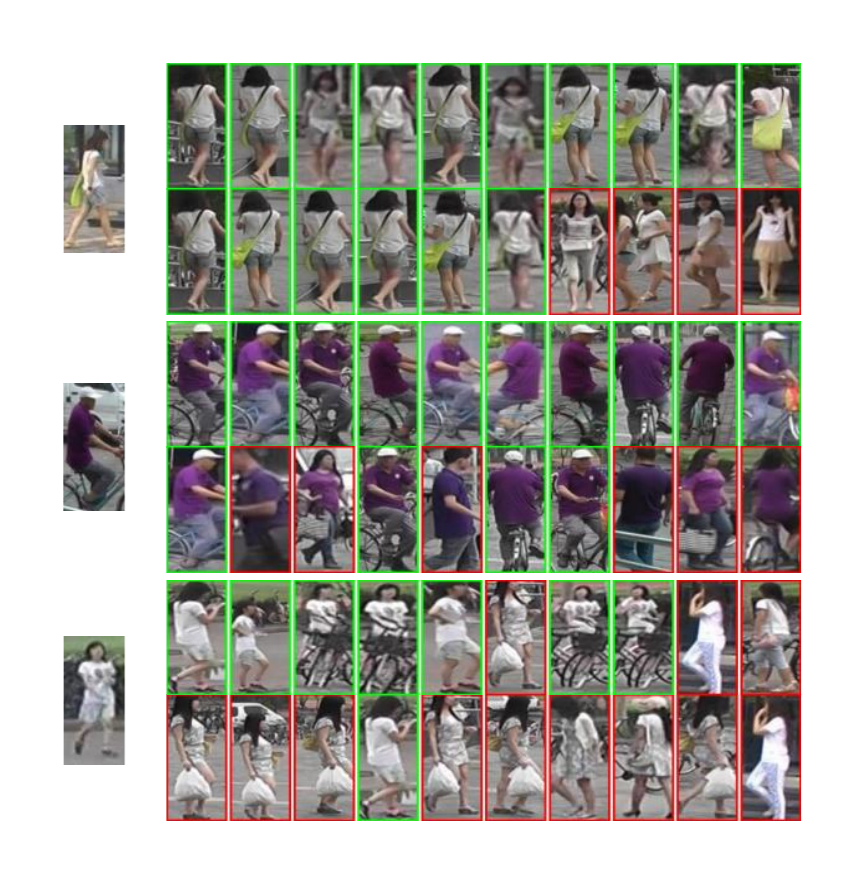}
\caption{Three Re-ID examples of PLR-OSNet and OSNet on DukeMTMC-reID. Left: query image. Upper-Right: top-10 results of PLR-OSNet. Low-Right: top-10 results of OSNet. Images in red boxes are negative results. PLR-OSNet boosts the retrieval performance}
\label{fig:visualDem}
\end{figure}

\textbf{Visualization of Feature Diversity Between Two Branches}: In Figure \ref{fig:demDetail}, we show the visualization of class activation maps (CAMs) for the global feature vector and 4 local part-level feature vectors. Note that the local branch produces 4 part-level feature vectors, corresponding to part-1, part-2, part-3 and part-4. As shown, these part-level features have some degree of diversity compared to the global features. This means that the proposed PLR-OSNet architecture allows the model to learn diverse features, which is key to the high performance of person Re-ID.

\textbf{Re-ID Visual Retrieving Results}:
We compare PLR-OSNet with OSNet more directly from visual retrieving results. Three retrieved examples are shown in Figure \ref{fig:visualDem}. One can see that OSNet fails to retrieve several correct images among the top-10 results.  Taking the second query as an example, PLR-OSNet is able to find correct images of the same identity in the top 10 results whilst OSNet gets 5 incorrect ones.

\subsection{Ablation Studies}

\subsubsection{Benefit of Global Features}
\begin{table*}[ht]
\begin{center}
\begin{tabular}{l|c@{\hskip 5pt}c@{\hskip 5pt}|c@{\hskip 5pt}c@{\hskip 5pt}|c@{\hskip 5pt}c@{\hskip 5pt}|c@{\hskip 5pt}c@{\hskip 5pt}}
\toprule[1.5pt]
\multirow{2}{*}{Global Features} &	\multicolumn{2}{c|}{Market1501}	 &	\multicolumn{2}{c|}{DukeMTMC} &	\multicolumn{2}{c|}{CUHK03-Labeled}	&			\multicolumn{2}{c}{CUHK03-Detected}			\\
\cline{2-9}
	&	mAP 	&	rank-1 &	mAP 	&	rank-1  &	mAP 	&	rank-1	&	mAP 	&	rank-1	\\
\hline\hline
No	&	86.9	&	94.6 &	79.8	&	90.2 &	77.5	&	81.2	& 73.4 & 77.6	\\
\hline
Yes &	\textbf{88.9} &	\textbf{95.6} & \textbf{81.2} &	\textbf{91.6} &	\textbf{80.5}	&	\textbf{84.6}	& \textbf{77.2}	& \textbf{80.4}	\\
\bottomrule[1.5pt]
\end{tabular}
\end{center}
\caption{The use of global features on the final performance}
\label{tb:global}
\end{table*}

\begin{table*}[ht]
\begin{center}
\begin{tabular}{l|c@{\hskip 5pt}c@{\hskip 5pt}|c@{\hskip 5pt}c@{\hskip 5pt}|c@{\hskip 5pt}c@{\hskip 5pt}|c@{\hskip 5pt}c@{\hskip 5pt}}
\toprule[1.5pt]
\multirow{2}{*}{Method} &	\multicolumn{2}{c|}{Market1501}	 &	\multicolumn{2}{c|}{DukeMTMC} &	\multicolumn{2}{c|}{CUHK03-Labeled}	&			\multicolumn{2}{c}{CUHK03-Detected}			\\
\cline{2-9}
	&	mAP 	&	rank-1 &	mAP 	&	rank-1  &	mAP 	&	rank-1	&	mAP 	&	rank-1	\\
\hline\hline
Multiple ID loss	&	85.6	&	94.4 &	77.0	&	89.4 &	79.4	&	83.1	& 74.7 & 78.4	\\
\hline
Single ID loss  &	\textbf{88.9} &	\textbf{95.6} & \textbf{81.2} &	\textbf{91.6} &	\textbf{80.5}	&	\textbf{84.6}	& \textbf{77.2}	& \textbf{80.4}	\\
\bottomrule[1.5pt]
\end{tabular}
\end{center}
\caption{Single ID Loss vs. Multiple ID Loss}
\label{tb:ID}
\end{table*}

\begin{table*}[ht]
\begin{center}
\begin{tabular}{l|c@{\hskip 5pt}c@{\hskip 5pt}|c@{\hskip 5pt}c@{\hskip 5pt}|c@{\hskip 5pt}c@{\hskip 5pt}|c@{\hskip 5pt}c@{\hskip 5pt}}
\toprule[1.5pt]
\multirow{2}{*}{Attention Modules} &	\multicolumn{2}{c|}{Market1501}	 &	\multicolumn{2}{c|}{DukeMTMC} &	\multicolumn{2}{c|}{CUHK03-Labeled}	&			\multicolumn{2}{c}{CUHK03-Detected}			\\
\cline{2-9}
	&	mAP 	&	rank-1 &	mAP 	&	rank-1  &	mAP 	&	rank-1	&	mAP 	&	rank-1	\\
\hline\hline
No	&	88.4	&	95.0 &	81.0	&	90.8 &	79.5	&	82.4	& 76.8 & 79.4	\\
\hline
Yes &	\textbf{88.9} &	\textbf{95.6} & \textbf{81.2} &	\textbf{91.6} &	\textbf{80.5}	&	\textbf{84.6}	& \textbf{77.2}	& \textbf{80.4}	\\
\bottomrule[1.5pt]
\end{tabular}
\end{center}
\caption{The use of attention modules on the final performance}
\label{tb:att}
\end{table*}
PCB employed a uniform partition strategy for producing part-level features, which did not consider any possibility of the use of global features. The proposed PLR-OSNet, however, introduces a global branch, which uses \textit{global-max-pooling} for extracting global features as shown in Figure \ref{fig:net}.  With the use of global features, PLR-OSNet performs significantly better as depicted in Table \ref{tb:global} for all the three datasets. For CUHK03-Label, PLR-OSNet achieves the Rank-1 accuracy of 84.6\% with the global features, without which it simply can obtain 81.2\% Rank-1 accuracy.   This suggests that the global branch and the local branch reinforce each other, both contributing to the final performance.

\subsubsection{Single ID Loss vs. Multiple ID Loss}
PLR-OSNet uses only single ID loss for multiple part-level features, which is sharply contrast to PCB and its variants, where each part-level feature vector is employed to drive an ID loss so that the number of ID loss is equal to the number of separated parts. The use of ID loss for each part-level feature can force it to learn the feature at each specified part with the ID-labeled dataset. The drawback, however, is that some part-level features may fail to produce any reliable ID prediction. By concatenating multiple part-level feature vector into a single feature vector, a single ID prediction is much more reliable.

With the use of multiple feature concatenation followed by a single ID loss, PLR-OSNet performs significantly better as shown in Table \ref{tb:ID} for all the three datasets.  For Market1501, PLR-OSNet obtains 88.9\% mAP, which surpasses its counterpart (with multiple ID loss) about 3.3\%.

\subsubsection{Benefit of Attention Modules}
The attention modules have been widely employed in various state-of-the-art methods for person Re-ID. Therefore, we also insert these popular attention modules in the shared net as shown in Figure \ref{fig:net}. Experiments results are shown in in Table \ref{tb:att} for all the three datasets. Clearly, it achieves consistently improved performance for all three datasets. However, the improvement is moderate, which may be due to the existence of the inherent attention mechanisms in OSNet.

\subsubsection{Soft Margin Triplet Loss vs. Hard Margin Triplet Loss}
\begin{table}
\begin{center}
\label{my-label}
\begin{tabular}{l|c@{\hskip 5pt}c@{\hskip 5pt}|c@{\hskip 5pt}c@{\hskip 5pt}}
\toprule[1.5pt]
\multirow{2}{*}{Triplet Loss}  &   \multicolumn{2}{c|}{CUHK03-Labeled}	&	\multicolumn{2}{c}{CUHK03-Detected}			\\
\cline{2-5}
	 &	mAP 	&	rank-1	&	mAP 	&	rank-1	\\
\hline\hline
Hard Margin &	75.9	&	78.6	& 72.8 & 76.0	\\
\hline
Soft Margin &	\textbf{80.5}	&	\textbf{84.6}	& \textbf{77.2}	& \textbf{80.4}	\\
\bottomrule[1.5pt]
\end{tabular}
\end{center}
\caption{Soft Margin Triplet Loss vs. Hard Margin Triplet Loss.}
\label{tb:softHard}
\end{table}

Table \ref{tb:softHard} studies the impact of soft margin triplet loss on the performance of the PLR-OSNet over CUHK03. Surprisingly, there is a large performance gap between soft margin triplet loss and hard margin triplet loss. We can see that the Rank-1 accuracy gap is around 6\% while the mAP gap is about 4.6\%. We also do experiments over Market1501 and DukeMTMC. However, experiments show that the use of soft margin triplet loss does not produce any observable improvement over the hard margin counterpart. Therefore, it remains unknown why the soft margin triplet loss can produce significantly better results compared to the hard version for CUHK03.

\section{Conclusion}
In this paper, we propose a new OSNet structure with part-level feature resolution for person Re-ID. With a two-branch network architecture, the proposed PLR-OSNet concatenates various uniformly-partitioned part-level feature vectors to a long vector for producing a single ID prediction loss, which is proved to be more efficient than the existing part-based methods.  Extensive experiments show that PLR-OSNet achieves state-of-the-art performance on popular person Re-ID datasets, including Market1501, DukeMTMC-reID and CUHK03. In the mean time, its model size is significantly smaller than various state-of-the-art methods, thanking to the lightweight architecture of OSNet.



\begin{thebibliography}{10}
\providecommand{\url}[1]{#1}
\csname url@samestyle\endcsname
\providecommand{\newblock}{\relax}
\providecommand{\bibinfo}[2]{#2}
\providecommand{\BIBentrySTDinterwordspacing}{\spaceskip=0pt\relax}
\providecommand{\BIBentryALTinterwordstretchfactor}{4}
\providecommand{\BIBentryALTinterwordspacing}{\spaceskip=\fontdimen2\font plus
\BIBentryALTinterwordstretchfactor\fontdimen3\font minus
  \fontdimen4\font\relax}
\providecommand{\BIBforeignlanguage}[2]{{%
\expandafter\ifx\csname l@#1\endcsname\relax
\typeout{** WARNING: IEEEtran.bst: No hyphenation pattern has been}%
\typeout{** loaded for the language `#1'. Using the pattern for}%
\typeout{** the default language instead.}%
\else
\language=\csname l@#1\endcsname
\fi
#2}}
\providecommand{\BIBdecl}{\relax}
\BIBdecl

\bibitem{zheng2016person}
{Zheng, Liang and Yang, Yi and Hauptmann, Alexander G. {(2016)}}, ``Person
  re-identification: Past, present and future,'' [Online]. Available:
  https://arxiv.org/abs/1610.02984.

\bibitem{dai2019BDB}
Z.~{Dai}, M.~{Chen}, X.~{Gu}, S.~{Zhu}, and P.~{Tan}, ``Batch dropblock network
  for person re-identification and beyond,'' in \emph{Proc. ICCV}, 2019, pp.
  3691--3701.

\bibitem{chen2019MHN}
B.~{Chen}, W.~{Deng}, and J.~{Hu}, ``Mixed high-order attention network for
  person re-identification,'' in \emph{Proc. ICCV}, 2019, pp. 371--381.

\bibitem{yang2019CAMA}
W.~{Yang}, H.~{Huang}, Z.~{Zhang}, X.~{Chen}, K.~{Huang}, and S.~{Zhang},
  ``Towards rich feature discovery with class activation maps augmentation for
  person re-identification,'' in \emph{Proc. CVPR}, June 2019, pp. 1389--1398.

\bibitem{hou2019IAN}
R.~{Hou}, B.~{Ma}, H.~{Chang}, X.~{Gu}, S.~{Shan}, and X.~{Chen},
  ``Interaction-and-aggregation network for person re-identification,'' in
  \emph{Proc. CVPR}, 2019, pp. 9317--9326.

\bibitem{chen2019ABD}
T.~{Chen}, S.~{Ding}, J.~{Xie}, Y.~{Yuan}, W.~{Chen}, Y.~{Yang}, Z.~{Ren}, and
  Z.~{Wang}, ``Abd-net: Attentive but diverse person re-identification,'' in
  \emph{Proc. ICCV}, 2019, pp. 8351--8361.

\bibitem{su2017pose}
C.~{Su}, J.~{Li}, S.~{Zhang}, J.~{Xing}, W.~{Gao}, and Q.~{Tian}, ``Pose-driven
  deep convolutional model for person re-identification,'' in \emph{Proc.
  ICCV}, 2017, pp. 3960--3969.

\bibitem{chen2017beyond}
W.~{Chen}, X.~{Chen}, J.~{Zhang}, and K.~{Huang}, ``Beyond triplet loss: A deep
  quadruplet network for person re-identification,'' in \emph{Proc. CVPR}, July
  2017, pp. 1320--1329.

\bibitem{bai2017reid}
S.~{Bai}, X.~{Bai}, and Q.~{Tian}, ``Scalable person re-identification on
  supervised smoothed manifold,'' in \emph{Proc. CVPR}, July 2017, pp.
  3356--3365.

\bibitem{he2019Bag}
H.~{He}, Z.~{Zhang}, H.~{Zhang}, Z.~{Zhang}, J.~{Xie}, and M.~{Li}, ``Bag of
  tricks for image classification with convolutional neural networks,'' in
  \emph{Proc. CVPR}, 2019, pp. 558--567.

\bibitem{zheng2017discriminatively}
Z.~Zheng, L.~Zheng, and Y.~Yang, ``A discriminatively learned cnn embedding for
  person reidentification,'' \emph{ACM Trans. Multimedia Comput., Commun.,
  Appl.}, vol.~14, no.~1, p.~13, 2018.

\bibitem{hermans2017defense}
{Hermans, Alexander and Beyer, Lucas and Leibe, Bastian. {(2017)}}, ``In
  defense of the triplet loss for person re-identification,'' [Online].
  Available: https://arxiv.org/abs/1703.07737.

\bibitem{su2016deep}
C.~Su, S.~Zhang, J.~Xing, W.~Gao, and Q.~Tian, ``Deep attributes driven
  multi-camera person re-identification,'' in \emph{Proc. ECCV}.\hskip 1em plus
  0.5em minus 0.4em\relax Springer, 2016, pp. 475--491.

\bibitem{yao2017deep}
H.~{Yao}, S.~{Zhang}, R.~{Hong}, Y.~{Zhang}, C.~{Xu}, and Q.~{Tian}, ``Deep
  representation learning with part loss for person re-identification,''
  \emph{IEEE Trans. Image Process.}, vol.~28, no.~6, pp. 2860--2871, June 2019.

\bibitem{sun2018beyond}
Y.~Sun, L.~Zheng, Y.~Yang, Q.~Tian, and S.~Wang, ``Beyond part models: Person
  retrieval with refined part pooling (and a strong convolutional baseline),''
  in \emph{Proc. ECCV}, 2018, pp. 480--496.

\bibitem{zhao2017deeply}
L.~{Zhao}, X.~{Li}, Y.~{Zhuang}, and J.~{Wang}, ``Deeply-learned part-aligned
  representations for person re-identification,'' in \emph{Proc. ICCV}, Oct
  2017, pp. 3239--3248.

\bibitem{dai2018batch}
{Dai, Zuozhuo and Chen, Mingqiang and Zhu, Siyu and Tan, Ping. {(2018)}},
  ``Batch feature erasing for person re-identification and beyond,'' [Online].
  Available: https://arxiv.org/abs/1811.07130.

\bibitem{kumar2017pose}
V.~{Kumar}, A.~{Namboodiri}, M.~{Paluri}, and C.~V. {Jawahar}, ``Pose-aware
  person recognition,'' in \emph{Proc. CVPR}, July 2017, pp. 6797--6806.

\bibitem{zheng2017pose}
L.~{Zheng}, Y.~{Huang}, H.~{Lu}, and Y.~{Yang}, ``Pose-invariant embedding for
  deep person re-identification,'' \emph{IEEE Trans. Image Process}, vol.~28,
  no.~9, pp. 4500--4509, Sep. 2019.

\bibitem{qian2018pose}
X.~Qian, Y.~Fu, T.~Xiang, W.~Wang, J.~Qiu, Y.~Wu, Y.-G. Jiang, and X.~Xue,
  ``Pose-normalized image generation for person re-identification,'' in
  \emph{Proc. ECCV}, 2018, pp. 650--667.

\bibitem{wang2018MGN}
G.~Wang, Y.~Yuan, X.~Chen, J.~Li, and X.~Zhou, ``Learning discriminative
  features with multiple granularities for person re-identification,'' in
  \emph{Proc. ACM Multimedia}, 2018, pp. 274--282.

\bibitem{suh2018part}
Y.~Suh, J.~Wang, S.~Tang, T.~Mei, and K.~Mu~Lee, ``Part-aligned bilinear
  representations for person re-identification,'' in \emph{Proc. ECCV}, 2018,
  pp. 402--419.

\bibitem{cheng2016person}
D.~{Cheng}, Y.~{Gong}, S.~{Zhou}, J.~{Wang}, and N.~{Zheng}, ``Person
  re-identification by multi-channel parts-based cnn with improved triplet loss
  function,'' in \emph{Proc. CVPR}, June 2016, pp. 1335--1344.

\bibitem{fan2018scpnet}
X.~Fan, H.~Luo, X.~Zhang, L.~He, C.~Zhang, and W.~Jiang, ``Scpnet:
  Spatial-channel parallelism network for joint holistic and partial person
  re-identification,'' in \emph{Proc. ACCV}.\hskip 1em plus 0.5em minus
  0.4em\relax Springer, 2018, pp. 19--34.

\bibitem{DeVries2017Cut}
{Terrance DeVries and Graham W Taylor. {(2018)}}, ``Improved regularization of
  convolutional neural networks with cutout,'' [Online]. Available:
  https://arxiv.org/abs/1708.04552.

\bibitem{zhong2017Erasing}
{Zhong, Zhun and Zheng, Liang and Kang, Guoliang and Li, Shaozi and Yang, Yi.
  {(2017)}}, ``Random erasing data augmentation,'' [Online]. Available:
  https://arxiv.org/abs/1708.04896.

\bibitem{zhou2019OSNet}
K.~Zhou, Y.~Yang, A.~Cavallaro, and T.~Xiang, ``Omni-scale feature learning for
  person re-identification,'' in \emph{Proc. ICCV}, 2019, pp. 3702--3712.

\bibitem{li2018harmonious}
W.~{Li}, X.~{Zhu}, and S.~{Gong}, ``Harmonious attention network for person
  re-identification,'' in \emph{Proc. CVPR}, June 2018, pp. 2285--2294.

\bibitem{wang2018mancs}
C.~Wang, Q.~Zhang, C.~Huang, W.~Liu, and X.~Wang, ``Mancs: A multi-task
  attentional network with curriculum sampling for person re-identification,''
  in \emph{Proc. ECCV}, 2018, pp. 365--381.

\bibitem{quan2019Auto}
R.~{Quan}, X.~{Dong}, Y.~{Wu}, L.~{Zhu}, and Y.~{Yang}, ``Auto-reid: Searching
  for a part-aware convnet for person re-identification,'' in \emph{Proc.
  ICCV}, 2019.

\bibitem{zheng2019pyramid}
F.~{Zheng}, C.~{Deng}, X.~{Sun}, X.~{Jiang}, X.~{Guo}, Z.~{Yu}, F.~{Huang}, and
  R.~{Ji}, ``Pyramidal person re-identification via multi-loss dynamic
  training,'' in \emph{Proc. CVPR}, June 2019.

\bibitem{Ristani2018Soft}
E.~{Ristani} and C.~{Tomasi}, ``Features for multi-target multi-camera tracking
  and re-identification,'' in \emph{Proc. CVPR}, June 2018, pp. 6036--6046.

\bibitem{chen2017multi}
W.~Chen, X.~Chen, J.~Zhang, and K.~Huang, ``A multi-task deep network for
  person re-identification,'' in \emph{Proc. AAAI}, 2017.

\bibitem{xia2019SONA}
B.~N. {Xia}, Y.~{Gong}, Y.~{Zhang}, and C.~{Poellabauer}, ``Second-order
  non-local attention networks for person re-identification,'' in \emph{Proc.
  ICCV}, 2019, pp. 3760--3769.

\bibitem{zhang2020Part}
Z.~Zhang and M.~Huang, ``Person re-identification based on heterogeneous
  part-based deep network in camera networks,'' \emph{IEEE Transactions on
  Emerging Topics in Computational Intelligence}, vol. Early Access, pp. 1--10,
  December 2018.

\bibitem{chang2018MFN}
X.~{Chang}, T.~M. {Hospedales}, and T.~{Xiang}, ``Multi-level factorisation net
  for person re-identification,'' in \emph{Proc. CVPR}, June 2018, pp.
  2109--2118.

\bibitem{qian2017mulscale}
X.~Qian, Y.~Fu, Y.-G. Jiang, T.~Xiang, and X.~Xue, ``Multi-scale deep learning
  architectures for person reidentification,'' in \emph{Proc. ICCV}, 2017.

\bibitem{chen2018person}
Y.~Chen, X.~Zhu, S.~Gong \emph{et~al.}, ``Person re-identification by deep
  learning multi-scale representations,'' in \emph{Proc. ICCV}, 2018.

\bibitem{hu2018SE}
J.~Hu, L.~Shen, and G.~Sun, ``Squeeze-and-excitation networks,'' in \emph{Proc.
  CVPR}, 2018, pp. 7132--7141.

\bibitem{wen2016discriminative}
Y.~Wen, K.~Zhang, Z.~Li, and Y.~Qiao, ``A discriminative feature learning
  approach for deep face recognition,'' in \emph{Proc. ECCV}.\hskip 1em plus
  0.5em minus 0.4em\relax Springer, 2016, pp. 499--515.

\bibitem{zheng2015scalable}
L.~{Zheng}, L.~{Shen}, L.~{Tian}, S.~{Wang}, J.~{Wang}, and Q.~{Tian},
  ``Scalable person re-identification: A benchmark,'' in \emph{Proc. ICCV}, Dec
  2015, pp. 1116--1124.

\bibitem{Ristani2016Performance}
E.~Ristani, F.~Solera, R.~Zou, R.~Cucchiara, and C.~Tomasi, ``Performance
  measures and a data set for multi-target, multi-camera tracking,'' in
  \emph{Proc. ECCV}.\hskip 1em plus 0.5em minus 0.4em\relax Springer, 2016, pp.
  17--35.

\bibitem{Li2014DeepReID}
W.~{Li}, R.~{Zhao}, T.~{Xiao}, and X.~{Wang}, ``Deepreid: Deep filter pairing
  neural network for person re-identification,'' in \emph{Proc. CVPR}, June
  2014, pp. 152--159.

\bibitem{zhong2017re}
Z.~{Zhong}, L.~{Zheng}, D.~{Cao}, and S.~{Li}, ``Re-ranking person
  re-identification with k-reciprocal encoding,'' in \emph{Proc. CVPR}, July
  2017, pp. 3652--3661.

\bibitem{saquib2018pose}
M.~S. {Sarfraz}, A.~{Schumann}, A.~{Eberle}, and R.~{Stiefelhagen}, ``A
  pose-sensitive embedding for person re-identification with expanded cross
  neighborhood re-ranking,'' in \emph{Proc. CVPR}, June 2018, pp. 420--429.

\bibitem{shen2018KPM}
Y.~{Shen}, T.~{Xiao}, H.~{Li}, S.~{Yi}, and X.~{Wang}, ``End-to-end deep
  kronecker-product matching for person re-identification,'' in \emph{Proc.
  CVPR}, June 2018, pp. 6886--6895.

\bibitem{chen2018CRF}
D.~{Chen}, D.~{Xu}, H.~{Li}, N.~{Sebe}, and X.~{Wang}, ``Group consistent
  similarity learning via deep crf for person re-identification,'' in
  \emph{Proc. CVPR}, June 2018, pp. 8649--8658.

\bibitem{li2018SNL}
K.~Li, Z.~Ding, K.~Li, Y.~Zhang, and Y.~Fu, ``Support neighbor loss for person
  re-identification,'' in \emph{Proc. ACM Multimedia}, 2018, pp. 1492--1500.

\bibitem{zeng2018HDLF}
M.~Zeng, C.~Tian, and Z.~Wu, ``Person re-identification with hierarchical deep
  learning feature and efficient xqda metric,'' in \emph{Proc. ACM Multimedia},
  2018, pp. 1838--1846.

\bibitem{yang2018LCNN}
J.~Yang, X.~Shen, X.~Tian, H.~Li, J.~Huang, and X.-S. Hua, ``Local
  convolutional neural networks for person re-identification,'' in \emph{Proc.
  ACM Multimedia}, 2018, pp. 1074--1082.

\bibitem{almazan2018re}
{Almazan, Jon and Gajic, Bojana and Murray, Naila and Larlus, Diane. {(2018)}},
  ``Re-id done right: towards good practices for person re-identification,''
  [Online]. Available: https://arxiv.org/abs/1610.02984.

\bibitem{wang2018resource}
Y.~{Wang}, L.~{Wang}, Y.~{You}, X.~{Zou}, V.~{Chen}, S.~{Li}, G.~{Huang},
  B.~{Hariharan}, and K.~Q. {Weinberger}, ``Resource aware person
  re-identification across multiple resolutions,'' in \emph{Proc. CVPR}, June
  2018, pp. 8042--8051.

\end{thebibliography}
\end{document}